\pdfoutput=1

\documentclass[11pt]{article}

\usepackage{EMNLP2023}
\usepackage{graphicx}      
\usepackage{amsthm,amsmath,amssymb}

\usepackage{longtable}
\usepackage{caption}
\usepackage{multirow}
\usepackage{array}
\usepackage{booktabs}
\usepackage{enumitem} 
\usepackage{tabularx}
\usepackage{mathrsfs}
\usepackage{makecell}
\usepackage{times}
\usepackage{latexsym}

\usepackage[T1]{fontenc}

\usepackage[utf8]{inputenc}

\usepackage{microtype}

\usepackage{inconsolata}

\title{Generating Diverse Training Samples for Relation Extraction with Large Language Models}

\author{Zexuan Li, Hongliang Dai$^{\ast}$, Piji Li \\
\textsuperscript{\rm 1} College of  Artificial Intelligence, \\
Nanjing University of Aeronautics and Astronautics, Nanjing, China\\
\textsuperscript{\rm 2} MIIT Key Laboratory of Pattern Analysis and Machine Intelligence, Nanjing, China\\
\textsuperscript{\rm 3} The Key Laboratory of Brain-Machine Intelligence Technology, Ministry of Education, Nanjing, China.\\
  \texttt{\{zexuanli, hongldai, pjli\}@nuaa.edu.cn} \\}

\begin{document}
\maketitle
\renewcommand{\thefootnote}{\fnsymbol{footnote}}
\footnotetext[1]{Corresponding author.}
\renewcommand{\thefootnote}{\arabic{footnote}}
\begin{abstract}
Using Large Language Models (LLMs) to generate training data can potentially be a preferable way to improve zero or few-shot NLP tasks. However, many problems remain to be investigated for this direction. For the task of Relation Extraction (RE), we find that samples generated by directly prompting LLMs may easily have high structural similarities with each other. They tend to use a limited variety of phrasing while expressing the relation between a pair of entities. Therefore, in this paper, we study how to effectively improve the diversity of the training samples generated with LLMs for RE, while also maintaining their correctness. We first try to make the LLMs produce dissimilar samples by directly giving instructions in In-Context Learning (ICL) prompts. Then, we propose an approach to fine-tune LLMs for diversity training sample generation through Direct Preference Optimization (DPO). Our experiments on commonly used RE datasets show that both attempts can improve the quality of the generated training data. We also find that comparing with directly performing RE with an LLM, training a non-LLM RE model with its generated samples may lead to better performance.
\end{abstract}

\section{Introduction}
Relation Extraction (RE) aims to identify and classify specific relation categories between pairs of entities from text. It is an important task in information extraction and has been deeply used in knowledge graph construction \citep{DBLP:journals/csur/ZhongWLPW24}, question and answer systems \citep{DBLP:conf/anlp/SrihariL00} and so on. Existing models \citep{DBLP:conf/www/ChenZXDYTHSC22,DBLP:conf/ijcnlp/ZhouC22,DBLP:conf/iclr/PaoliniAKMAASXS21} applied to RE have achieved good results on many benchmarks. However, since there are various types of entity relations in different domains, data scarcity is a common problem while developing RE models in practice. Existing studies typically address this problem with techniques such as prototypical networks \cite{DBLP:conf/naacl/LiuHWC22}, meta-learning \cite{DBLP:conf/icml/QuGXT20} and prompt-tuning \cite{DBLP:conf/www/ChenZXDYTHSC22} under zero or few-shot settings.

\begin{figure}[t]
\centering
\centerline{\includegraphics[scale=0.24]{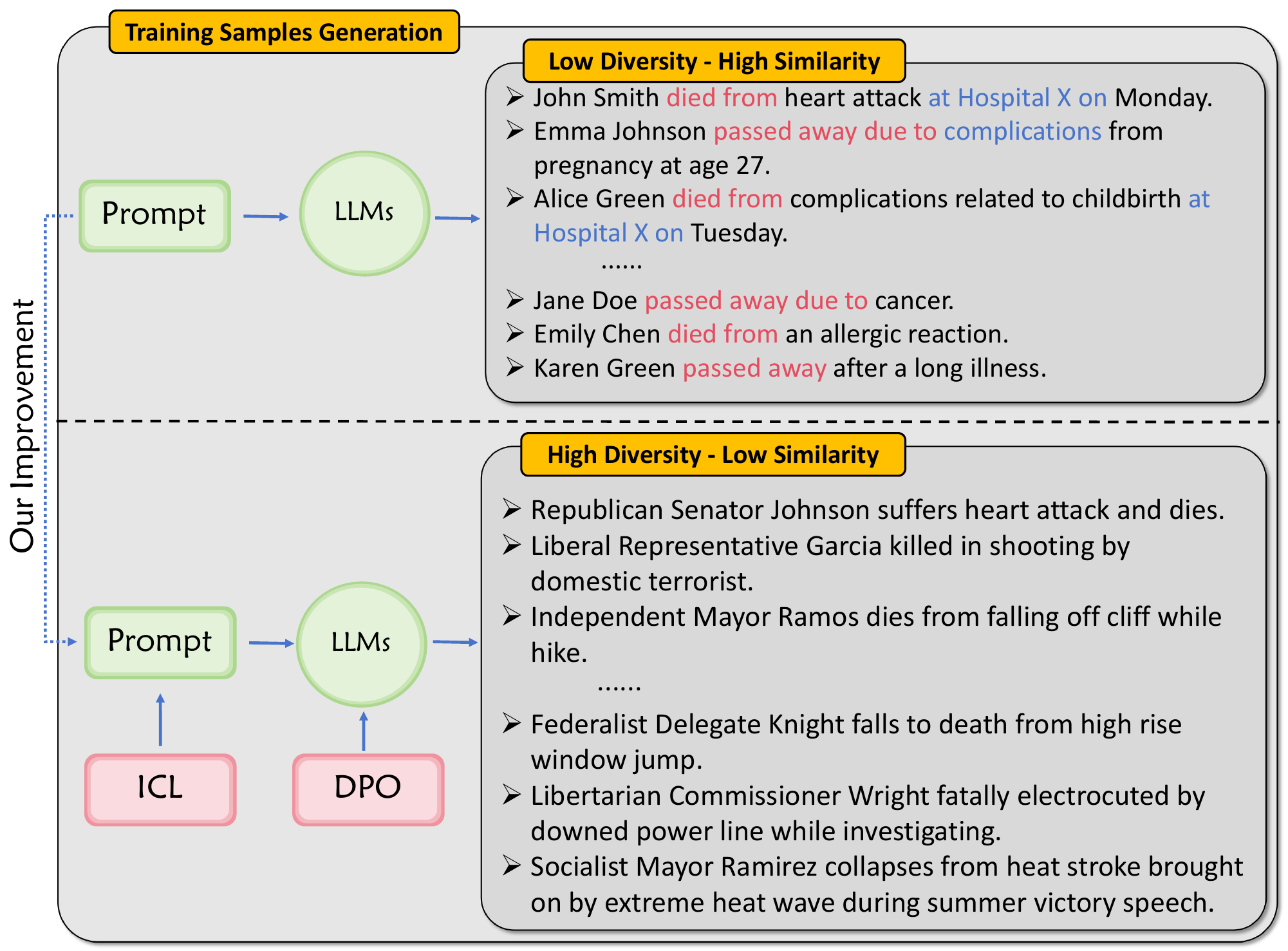}}
\caption{Training samples generated by LLMs for RE before and after adopting our approach.}
\label{fig:1}
\end{figure}

Recently, the powerful generative capabilities of Large Language Models (LLMs) \citep{OpenAI2022,OpenAI2023,DBLP:conf/nips/BrownMRSKDNSSAA20} have made it possible to alleviate the data scarcity problem with a new way: using LLMs to generate extra training data for the task. In this paper, we study the application of this approach to relation extraction. We find that directly prompting LLMs to obtain RE training samples may easily lead to low diversity results. An example is shown in the upper part of Figure \ref{fig:1}, the LLM repeatedly uses a limited number of key verbs or phrases to indicate the relation between the head and tail entities. The structures of the sentences are also almost the same.

We therefore focus on how to improve the diversity of LLM generated training instances for RE, while maintaining the correctness. To this end, several approaches are proposed and tried. First, we employ the In-Context Learning (ICL) \citep{DBLP:conf/emnlp/WanCMLSLK23,DBLP:journals/corr/abs-2301-00234,DBLP:conf/nips/BrownMRSKDNSSAA20} technique. The most straightforward way to improve diversity through ICL is to directly give the model an extra instruction in the prompt, asking it to generate dissimilar samples. In addition, we also propose a one by one generation procedure, where each time, the model is instructed to generate one more training sample that is different from the given demonstrations.

Conducting ICL cannot change the inherent behavior of LLMs. Thus, we also propose an approach to fine-tune LLMs for diversity RE sample generation. We adopt Direct Preference Optimization (DPO) for fine-tuning, which has been verified to perform well on many tasks such as summary generation and single-round conversations \citep{DBLP:conf/nips/RafailovSMMEF23}. RE samples that are similar with existing ones and incorrect samples are automatically generated to serve as dispreferred answers to the DPO algorithm, thus training the LLM to consider both diversity and correctness. The lower part of Figure \ref{fig:1} is an example of the training instances obtained through our approach. It can be seen that these samples generated by the LLMs are more diverse in terms of the verbs or phrases that indicate relations, and the overall architectures of the sentences.

Finally, we conduct extensive experiments on commonly used RE datasets including three versions of TACRED and SemEval to evaluate the effectiveness of our method. We also show that comparing with directly performing RE with an LLM, training a non-LLM RE model with its generated samples can potentially lead to better performance.

Our main contributions as summarized follows:

\begin{itemize}
    \item  We investigate ICL-based methods for diverse RE training sample generation with two different procedures: \textit{one by one} and \textit{all at once}.
    \item We propose an approach to fine-tuning LLMs with DPO that aims for generating diverse and correct RE training samples. The approach uses automatically constructed dispreferred answers, therefore reduces the requirement of human annotation.
    \item We provide comprehensive experimental results to analyze the performance of the proposed methods.
\end{itemize}
Our code is available at \href{https://github.com/Lzx-ZBC}{https://github.com/Lzx-ZBC}.

\begin{figure*}[t]
\centering
\centerline{\includegraphics[scale=0.4]{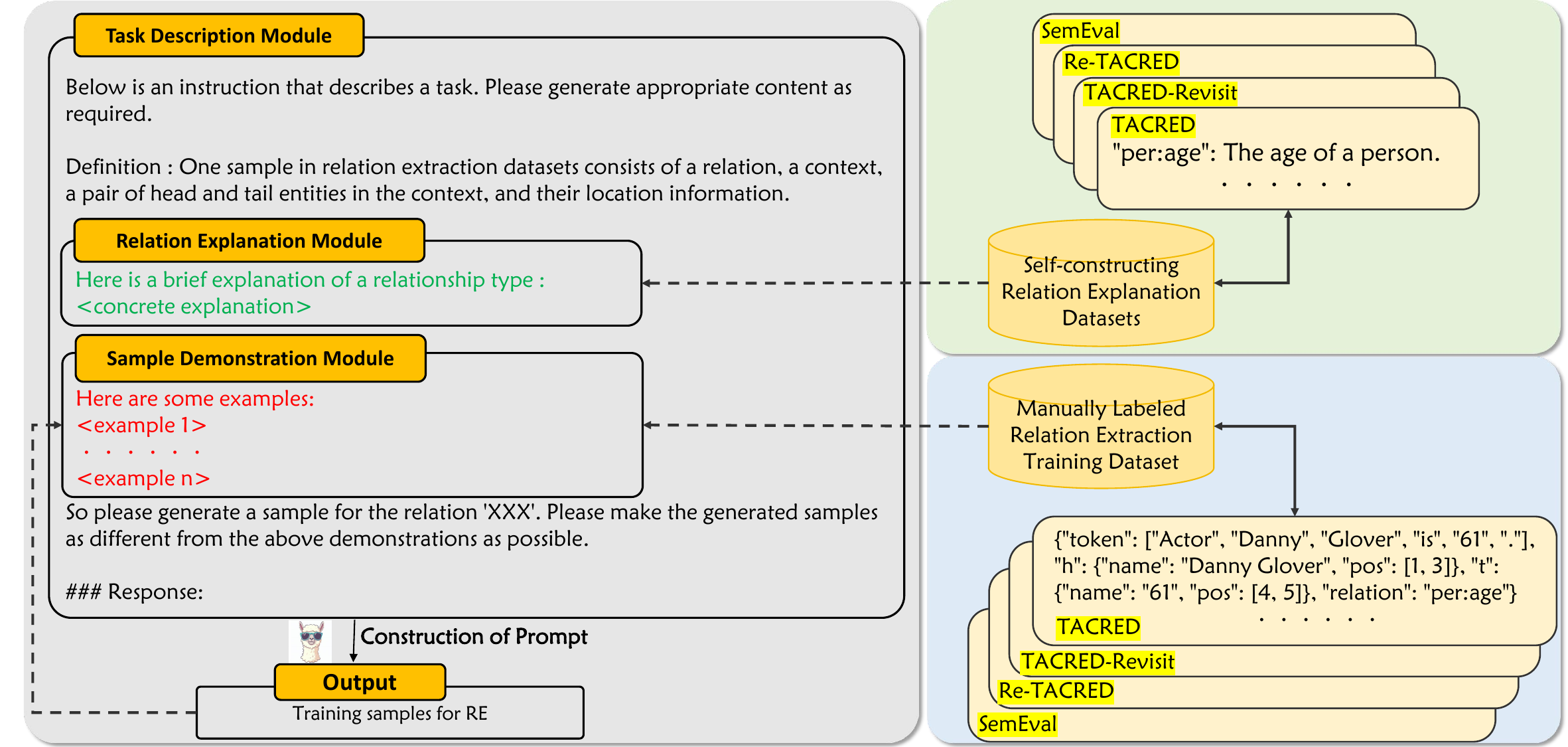}}
\caption{Construction of Prompt, which consists of three modules: Task Description Module, Relation Explanation Module and Sample Demonstration Module.}
\label{fig:prompt}
\end{figure*}

\begin{figure*}[t]
\centering
\centerline{\includegraphics[scale=0.468]{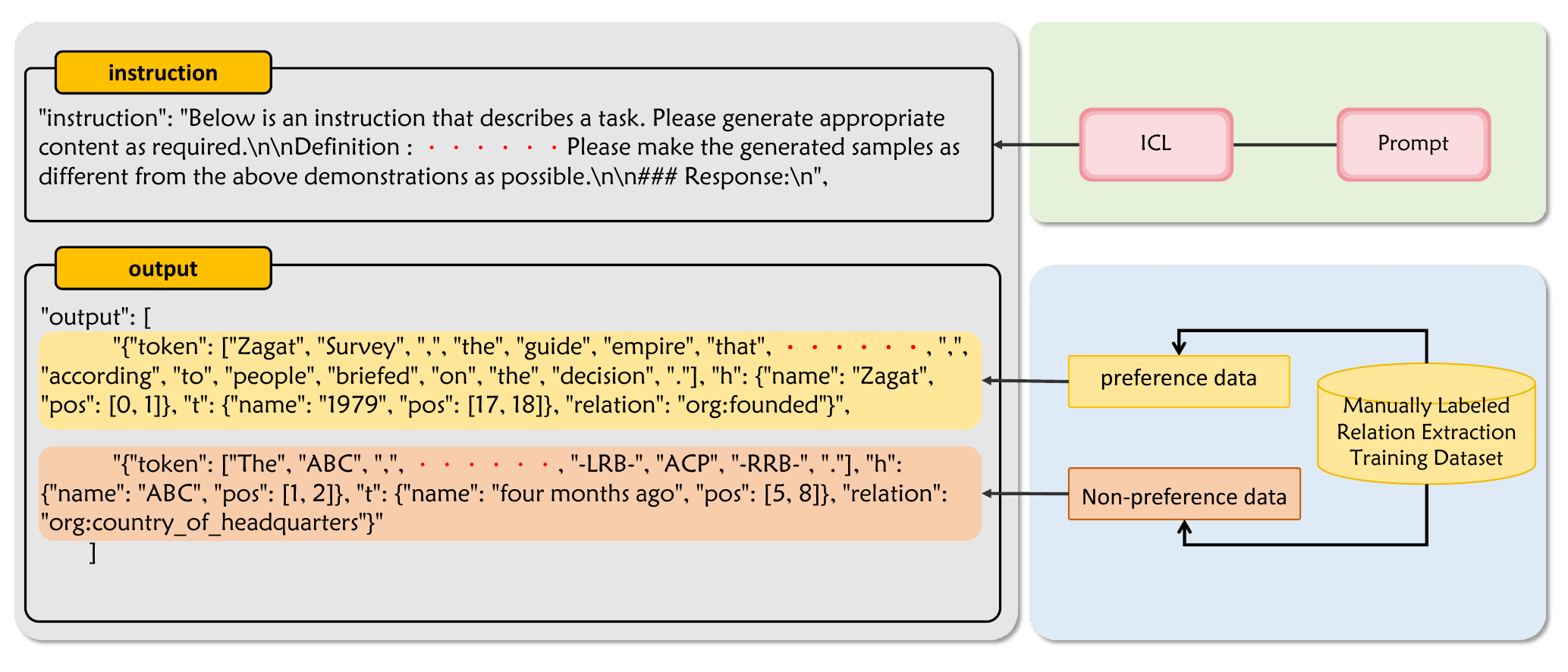}}
\caption{Construction of the DPO Fine-Tuning Training Dataset.}
\label{fig:DPO}
\end{figure*}

\section{Related Work}

\subsection{Relation Extraction with LLMs}
Relation Extraction (RE) aims to extract the relationship between head and tail entities based on their relevant context. The task can be approached by using traditional neural network models such as CNN and RNN \cite{DBLP:conf/emnlp/ZengLC015,DBLP:conf/emnlp/ZhangZCAM17,DBLP:conf/acl/ZhouSTQLHX16}. Since the proposal of Pre-Trained Language Models (PLMs) like BERT \cite{DBLP:conf/naacl/DevlinCLT19}, PLMs-based RE models has become the main solution due to their preferable performance. Alternatively, graph neural networks can also be employed for conducting RE \cite{DBLP:conf/ijcai/GuoN0C20,DBLP:conf/acl/GuoZL19,DBLP:conf/emnlp/Zhang0M18}. 

Recently, there is an increasing interest in using LLMs to perform RE directly. \citet{DBLP:conf/sustainlp/XuZWZ23} put a list of all relation categories and concrete samples in the prompt to make LLMs understand the exact process of RE. Their experimental results show that LLMs is capable of producing high quality RE predictions. \citet{DBLP:conf/acl/ZhangG023} propose a framework called QA4RE that coordinates RE with question and answer (QA). \citet{DBLP:conf/emnlp/LiWK23} proposes a new prompting method, SUMASK, which converts the input into a valid QA format using LLMs by decomposing RE into text summarization and QA.

\subsection{Data Generation with LLMs}

Data generation is gradually becoming a new focus topic. There were already some studies on this topic before instruction tuned LLMs become popular \cite{DBLP:conf/nips/MengHZH22,DBLP:conf/emnlp/YeGLXF00K22,DBLP:conf/iclr/GaoPLXY0ZLLK23}. For example, \citet{DBLP:conf/nips/MengHZH22} generates training data for NLU tasks such as sentiment classification through prompting PLMs. With instruction tuned LLMs, data generation becomes more convenient. \citet{DBLP:conf/acl/ChiaBPS22} proposes a framework to synthesize unseen relation types by prompting language models to generate structured text. \citet{DBLP:conf/sustainlp/XuZWZ23} use LLMs to generate data to assist the models themselves on RE. They accomplish this by describing the data content and samples in detail in the prompt so that LLMs can generate reasonable data. Recently, as instruction tuned LLMs have entered the research field, generating instruction tuned datasets requires careful writing of instructions and input-output pairs, which are usually written by humans, smaller in size and less in diversity. To overcome this problem, self-instruct \citep{DBLP:conf/acl/WangKMLSKH23} proposes a method for generating instruction tuned datasets by prompting available LLMs. The results show that the data generated by LLMs can improve their classification ability on RE. However, they do not put a specific focus on how the diversity and correctness of the generated training samples can be improved.

\section{Methodology}
In this section, we first introduce how we generate training samples for RE by directly performing in-context learning with LLMs. Then, we describe our approch to fine-tune LLMs for diverse and correct sample generation with DPO.
\subsection{Sample Generation with ICL}
\label{sub:ICL}

To apply in-context learning (ICL), we provide sample demonstrations for LLMs in the prompt to stimulate their understanding of the training instances for the relation extraction task.
With the demonstrations, the straightforward way to prompt LLMs is to instruct them to directly generate the training samples \textit{all at once}.
Here, in addition, we adopt an alternative way. We generate training samples \textit{one by one} instead of \textit{all at once}. Doing so enables the training samples generated by the LLMs to be added back to the prompt,
allowing us to instruct the model not to generate new samples similar with the already existing ones. Next, we introduce this method in detail. The method of generating \textit{all at once} will be explained in the experimental Results and Analysis section.
The prompt for LLMs constructed by us are shown in Figure \ref{fig:prompt}. It includes three modules: task description, relation explanation and sample demonstration. Their specific contents are as follows:

\paragraph{Task Description Module.} 
The Task Description Module covers the beginning and the end of the full prompt.
At the beginning of the prompt, we tell LLMs that it will accomplish the task of generating training samples for relation extraction, and detail that a sentence-level training sample for RE includes the key information of text, head and tail entities with their locations, and the relation category. At the end of the prompt, we order LLMs to generate new training samples for the relation category described in the Relation Explanation Module, and require that the generated training samples be as different as possible from the provided demonstrations in the Sample Demonstration Module.

\paragraph{Relation Explanation Module.} We construct a corresponding explanation for each relation category in the relation extraction dataset, and then put them into the Relation Explanation Module to allow the LLMs to further understand the relation type.
The details of the explanations for all relation categories are provided in Appendix \ref{sub:RES}.

\paragraph{Sample Demonstration Module.} For a specific relation category, we randomly select a manually labeled sample from the existing relation extraction dataset and put it into the Sample Demonstration Module for LLMs' reference, so that LLMs can further understand the structure and content of the relation extraction training samples that need to be generated. Once LLMs have completed one generation, the generated training instance is moved back to the Sample Demonstration Module to form a new Prompt. Under cyclic generation, the number of demonstration samples in this module will increase one by one as the number of generated samples increases. 

\subsection{Diversity Fine-tuning with DPO}
\label{sub:DPO}

Direct Preference Optimization (DPO) is an automated fine-tuning method that optimizes model parameters by maximizing the rewards of a pre-trained model on a specific task. Compared to traditional fine-tuning methods, DPO bypasses the step of modeling the reward function and instead improves performance by optimizing the model directly on the preference data. Given a human preference dataset $\mathcal{D}=\{(x_i,y_{i,1},y_{i,2})\}$, where $y_{i,1}$ is preference data, $y_{i,2}$ is non-preference data, the following objective can be optimized:
\begin{equation}
\label{eq:dpo}
    \max_{\pi}\Sigma_i\log\sigma\left(\frac{1}{\beta}\log\frac{\pi(y_{i,1}|x_i)}{\pi(y_{i,2}|x_i)}\right),
\end{equation}
where $\pi(y|x)$ is the optimization strategy. According to this equation, the key to DPO fine-tuning is to construct suitable preference and non-preference data for LLMs to make comparisons.

\paragraph{Construction of the DPO Fine-Tuning Data.} 
Each training instance for DPO consists of an input prompt ("instruction"), as well as two responses ("output"): a preferred response and an non-preferred response. As shown in Figure 3, we use the previously described ICL prompt as "instruction". Both the preference and non-preference data will be placed in the "Output". The preference data is a random sample of the target relation category directly extracted from the manually labeled dataset. The non-preference data is constructed in the following three ways: (1) Use a manually labeled training sample that belongs to another relation category, and modify its label to the target relation category. This constructs an sample that is incorrectly annotated as the target relation. (2) Choose a demonstration sample in the ``Instruction'' item, then replace the head and tail entities, and add or delete a few words in the context. This creates an instance that is similar to one of the demonstration samples. (3) Use a sample that is exactly the same as one of the demonstration samples in the ``Instruction'' item. 

Here, note that since we focus on the data scarcity scenario, there won't be enough manually annotated data of the target relation types for DPO fine-tuning. Thus, the source data for DPO should come from other already existing datasets that probably use different relation types or are of different domains.
In our experiments, to mimic such a scenario, we split the datasets by relation categories and use different parts for DPO and sample generation, respectively. This will be detailed in the Experimental Settings section.

\begin{figure}[h]
\centering
\centerline{\includegraphics[scale=0.355]{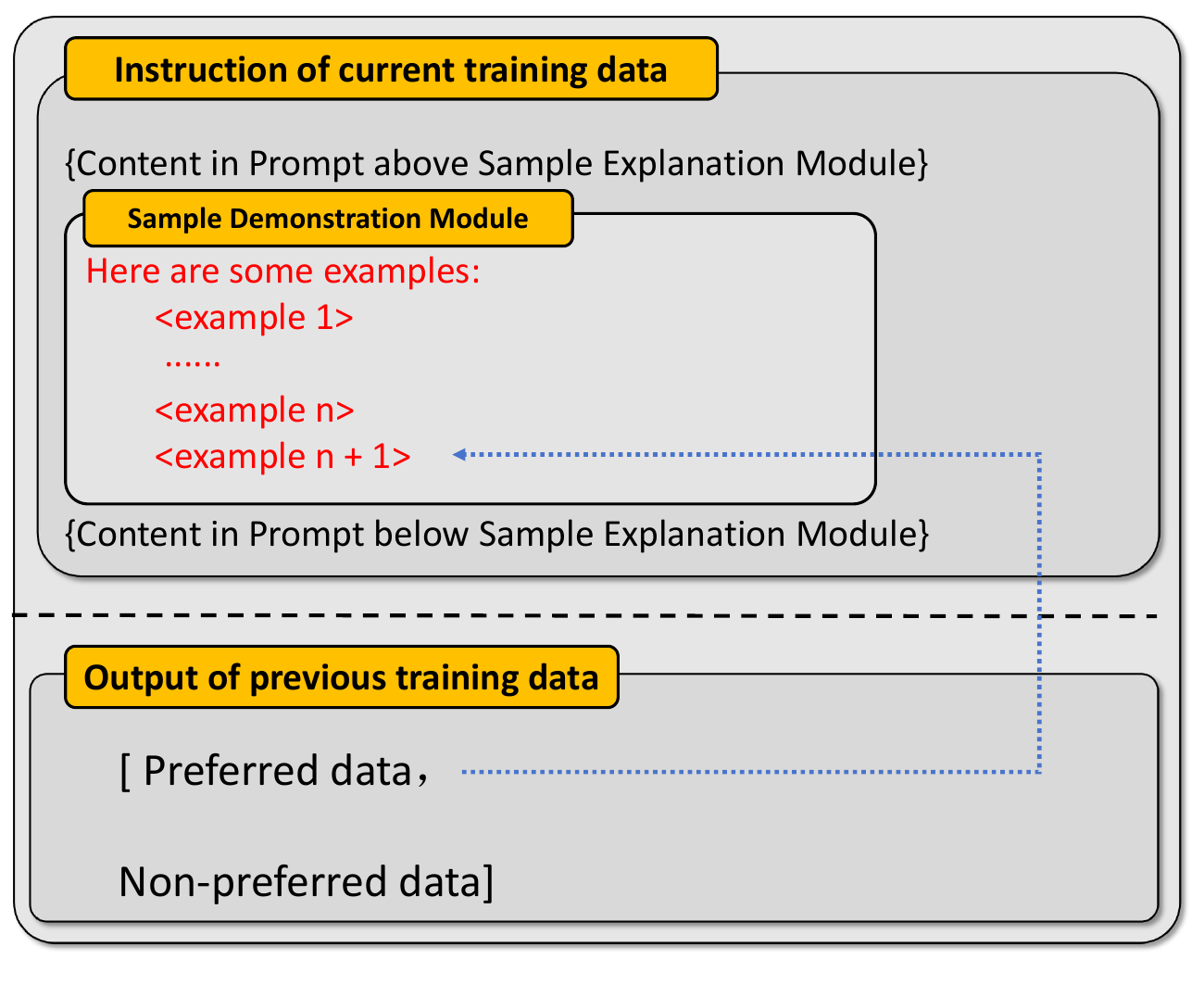}}
\caption{Imitating one by one generation during Direct Preference Optimization.}
\label{fig:adjust}
\end{figure}

\paragraph{Imitating One by One Generation.} 
As shown in Figure \ref{fig:adjust}, in order for the DPO fine-tuning to reflect the one by one generation process, for each relation category, we let the number of demonstration samples in the ``Instruction'' item of the DPO fine-tuning training dataset increment, specifically by placing the preferred manually labeled training samples from the ``Output'' item of the previous training data into the Sample Demonstration Module of the current training data while forming the ``Instruction'' item of new training data.

\section{Experimental Settings}
\subsection{Datasets}
We conduct experiments on SemEval and three versions of TACRED: SemEval 2010 Task 8 (SemEval) \citep{DBLP:conf/semeval/HendrickxKKNSPP10}, TACRED \citep{DBLP:conf/emnlp/ZhangZCAM17}, TACRED-Revisit \citep{DBLP:conf/acl/AltGH20a}, Re-TACRED \citep{DBLP:conf/aaai/StoicaPP21}. Statistical details are given in Table \ref{tab:dataset} and Appendix \ref{sub:datasets}.

\begin{table}[h]
\centering
\renewcommand{\arraystretch}{1}
\scalebox{0.8}{
\begin{tabular}{lrrrc}
\hline
Dataset & Train& Val & Test &Relation\\
\hline
SemEval & {6,507}& {1,493}& {2,717}& {19} \\ 
TACRED & {68,124} & {22,631}& {15,509}& {42}\\
TACRED-Revisit & {68,124}& {22,631}& {15,509}& {42} \\
Re-TACRED & {58,465}& {19,584}& {13,418}& {40} \\ 
\hline
\end{tabular}}
\caption{Statistics of the RE datasets. Including the numbers of instances in different splits and the numbers of relations.}
\label{tab:dataset}
\end{table}

\subsection{Evaluation}
Considering that it is difficult to directly assess the quality of the generated training samples, we put the generated training samples into the KnowPrompt \citep{DBLP:conf/www/ChenZXDYTHSC22} and RetrievalRE \citep{DBLP:conf/sigir/ChenLZTHSC22} for training. KnowPrompt achieves satisfying performance by Knowledge Injection and Synergistic Optimization. RetrievalRE, on the other hand, improves the generalization ability of the model when dealing with difficult patterns by combining retrieval enhancement and prompt tuning. Their performance on the test dataset reflects the quality of the generated training samples. The better the performance of them, the higher the quality of the generated training samples.

Finally, we follow existing RE studies and adopt Micro F1 as the evaluation metric.

\subsection{Implementation Details}

For KnowPrompt and RetrievalRE, we follow \cite{DBLP:conf/www/ChenZXDYTHSC22,DBLP:conf/sigir/ChenLZTHSC22} and use RoBERTA\_LARGE \cite{DBLP:journals/corr/abs-1907-11692} in all the experiments for a fair comparison. For LLMs, considering the cost and fine-tuning requirements, we used LLaMA2-7b-Chat \citep{DBLP:journals/corr/abs-2307-09288} in our experiments. We set temperature = 0.4, top\_p = 0.9, top\_k = 20, repetition\_penalty = 1.15. Meanwhile, we use LoRA to accomplish DPO fine-tuning of LLMs. We set truncation length = 1024, learning rate = 5e-5, batch size = 4, epoch = 20.

In order to prevent the risk of ``cheating'' on LLMs generation caused by DPO fine-tuning, we separated the relation categories used for DPO fine-tuning from those to be generated. 
Specifically, for each dataset, we divide the relation categories in half, thereby also partition the dataset into two parts. Then one part is used for DPO fine-tuning, and the other part is used for sample generation. By swapping the two parts, we are able to complete the sample generation of all relation categories.

\begin{table*}[h]
\centering
\renewcommand{\arraystretch}{1}
\scalebox{0.86}{
\begin{tabular}{*{13}{c}}
  \toprule
  \multirow{2}*{Method} & \multicolumn{3}{c}{TACRED} & \multicolumn{3}{c}{TACRED-Revisit} & \multicolumn{3}{c}{Re-TACRED} & \multicolumn{3}{c}{SemEval} \\
  \cmidrule(lr){2-13}
  & K=8 & K=16 & K=32 & K=8 & K=16 & K=32 & K=8 & K=16 & K=32 & K=8 & K=16 & K=32 \\
  \midrule
  Direct-RE & \multicolumn{3}{c}{21.17} & \multicolumn{3}{c}{21.82} & \multicolumn{3}{c}{33.30} & \multicolumn{3}{c}{25.55} \\
  \midrule
  Data Generation & 21.67 & 24.71 & 26.19 & 23.14 & 27.89 & 29.80 & 28.52 & 32.21 & 32.04 & - & - & - \\
  KnowPrompt & 22.07 & 30.00 & \textbf{36.33} & 26.31 & 30.76 & 34.61 & 44.77 & 56.51 & 62.34 & 61.39 & 74.08 & \textbf{81.06} \\ 
  Ours (pure) & 22.48 & 27.99 & 30.41 & 23.04 & 29.25 & 31.12 & 34.77 & 40.33 & 50.72 & 42.02 & 45.05& 47.70 \\	
  Ours (mix-OBO) & 33.48 & 34.93& 36.25 & \textbf{34.33} &34.39 & 35.27 & \textbf{57.41}&\textbf{61.09}& \textbf{64.58} & 67.92& 75.36& 80.40 \\
  Ours (mix-AAO) & \textbf{34.30} & \textbf{35.56}& 35.66 & 34.15 & \textbf{34.77} & \textbf{35.65} & 56.01&59.97& 63.55 & \textbf{70.32}& \textbf{76.36}& 80.46 \\  
  Ours (constant)&17.39 &21.71& - & 18.39&22.22& - & 36.01& 39.53 & - & 22.32&30.53& - \\
  \midrule
  Data Generation & 24.97 & 30.54 & 25.99 & 27.67 & 28.46 & 29.48 & 31.21 & 35.03 & 35.94 & - & - & - \\
  RetrievalRE & 32.04&35.16	&37.07 & 28.95&32.34&\textbf{37.26} & 33.73	&55.79&61.80 & 72.53&\textbf{81.42}& 83.93\\  
  Ours (pure) & 25.20 & 29.63 & 30.29 & 28.55 & 31.73 & 31.51 & 27.92 & 39.21 & 50.05 & 45.13 & 47.20& 49.22 \\	
  Ours (mix-OBO) & 33.75 & 35.83& \textbf{37.46} & 33.81 & 34.15 & 36.90 & \textbf{58.60}&\textbf{62.61}& \textbf{65.15} & \textbf{74.20}& 80.84& \textbf{84.01} \\
  Ours (mix-AAO) & \textbf{34.91} & \textbf{36.57}& 37.23 & \textbf{35.06}& \textbf{35.64}& 36.77 &55.43&61.15& 64.85 & 73.36& 78.90& 81.22 \\
  Ours (constant)&22.48 &25.31& - & 23.99&26.38& - & 33.28& 41.48 & - & 36.65&39.47& - \\
  \bottomrule
\end{tabular}}
\caption{\label{main-result}
Micro F1 (\%) of few-shot performance. \textbf{Direct-RE} means using LLMs directly for RE. \textbf{KnowPrompt} and \textbf{RetrievalRE} means the performance of manually labeled training samples. \textbf{Ours (pure)} means the performance of using LLM-generated training samples only. \textbf{Ours (mix-OBO)} and \textbf{Ours (mix-AAO)} mean the performance of combining the use of manually labeled training samples and training samples generated by LLM based on the OBO and AAO. \textbf{Ours (constant)} means the performance of LLMs-generated training samples with a fixed number of demonstration samples in Prompt. \textbf{Data Generation} means the performance of training samples generated by \citep{DBLP:conf/sustainlp/XuZWZ23}.
}
\end{table*}

\section{Results and Analysis}
\subsection{Main Results}

We compare with directly conducting RE using the LLM through ICL, and directly training KnowPrompt and RetrievalRE using only manually labeled data. The results are in Table \ref{main-result}. For Direct-RE, 2 samples are used as demonstrations. For our methods, 64 samples are generated for training. Note that in practice, more generated samples can be used if the performance can be further improved. An analysis over the number of generated samples is provided in Section 5.4.

\paragraph{Comparing to Manually Labeled Training Samples.}
In Table \ref{main-result}, Ours (pure) only uses the LLM-generated samples for training, while KnowPrompt and RetrievalRE only uses the manually labeled samples. 
The performance of Ours (pure) is comparable to that of KnowPrompt and RetrievalRE on TACRED and TACRED-Revisit, but is much worse on Re-TACRED and SemEval. Therefore, the qualities of the generated samples are good enough to be beneficial for training. But they still cannot be used to for the purpose of fully replacing manually labeled data, even when the number of training samples is as small as 8.
\paragraph{Comparing to Generated Training Sample by Another Way.} As demonstrated in Table \ref{main-result}, Ours (Pure) performs the baseline (Data Generation) across most experimental settings on the three variants of TACRED, indicating the superior efficacy of our sample generation methodology.

\paragraph{Comparing to Performing Relation Extraction Directly with LLMs.} 

As shown in Table \ref{main-result}, Ours (pure) performs better than directly using LLM to conduct relation extraction (Direct-RE) on all four relation extraction datasets. This means that when people employ LLMs for RE without any manually labeled data, they can consider training a non-LLM model with LLM-generated samples, instead of directly prompting the LLMs. This would also reduce the cost of repeatedly calling LLMs. Moreover, the self-improvements of LLMs on the RE task can also be an interesting direction.



\paragraph{Comparing to Mixed Training Samples.} We mix manually labeled training samples with LLMs-generated training samples and use them for the training of KnowPrompt and RetrievalRE. This corresponds to Ours (mix-OBO, mix-AAO) in Table \ref{main-result}, where K denotes the number of manually labeled training samples. The number of LLMs-generated training samples varies dynamically, and we choose the value in [8, 16, 32] that gives the best performance for KnowPrompt and RetrievalRE. We find that the model's performance is higher than when only pure LLM-generated training samples or only pure manually labeled training samples are used under most of the settings. This shows that the generated samples can be combined with existing human annotated data to help improve the final performance of RE models. However, as the amount of human annotated training data increases, the generated samples become less beneficial. 

\begin{table*}[t]
\centering
\renewcommand{\arraystretch}{1}
\scalebox{0.77}{
\begin{tabular}{*{13}{c}}
  \toprule
  \multirow{2}*{Method} & \multicolumn{3}{c}{TACRED} & \multicolumn{3}{c}{TACRED-Revisit} & \multicolumn{3}{c}{Re-TACRED} & \multicolumn{3}{c}{SemEval} \\
  \cmidrule(lr){2-13}
  & K=8 & K=16 & K=32 & K=8 & K=16 & K=32 & K=8 & K=16 & K=32 & K=8 & K=16 & K=32 \\
  \midrule
  Ours (AAO) & \textbf{21.86} & 26.91 & 26.39 & \textbf{23.17} & 27.97 & 28.50 & \textbf{38.00} & \textbf{45.68} & \textbf{46.50} & \textbf{41.49}& \textbf{54.53} & \textbf{56.34} \\
  Ours (AAO), w/o DPO & 18.97 & \textbf{27.70} & \textbf{36.04} & 22.61 & \textbf{31.08} & \textbf{37.08} & 20.64 & 36.68 & 36.66 & 37.66 & 47.49& 49.58 \\
  \midrule
  Ours (OBO) & \textbf{22.48} & \textbf{27.99} & \textbf{30.41} & \textbf{23.04} & \textbf{29.25} & \textbf{31.12} & \textbf{34.77} & 40.33 & \textbf{50.72} & \textbf{42.02} & \textbf{45.05} & \textbf{47.70} \\
  Ours (OBO), w/o DPO & 18.35 & 21.47 & 23.70 & 18.39 & 24.16& 27.52  & 31.82 & \textbf{41.47} & 40.45 & 24.65 & 28.42 & 30.03 \\
  Ours (OBO), w/o DPO, w/o DI & 21.86 & 21.75 & 25.66 & 21.90 & 24.11 & 27.90 & 33.50 & 33.55 & 32.74 & 20.97 & 27.01 & 25.88 \\
  \bottomrule
\end{tabular}
}
\caption{\label{ablation}
Micro F1 (\%) of different variants of our approach on KnowPrompt. \textbf{AAO} means generating all training samples at once. \textbf{OBO} means generating training samples one by one. \textbf{DI} means the instruction about ``diversity''. The best results in each column are indicated by \textbf{bolding}.
}
\end{table*}

\paragraph{Comparing to Using a Constant Number of Demonstration Samples.} We fixed the number of samples in the Sample Demonstration Module to 4. When the LLMs generate a new training sample, and if the number of samples demonstrated in the Sample Demonstration Module does not reach 4, the newly generated samples are directly added to this Module. Conversely, if the number reaches 4, we let it randomly replace one of the samples in the Sample Demonstration Module. As shown in Table \ref{main-result}, after the training samples generated under this approach are used for KnowPrompt or RetrievalRE training, the performance of the models drop substantially. We found by tracking the generation process of LLMs that when a training sample in the Sample Demonstration Module is replaced, the LLMs have a high probability of generating instances similar or even identival to this sample in the subsequent period, thus leading to a poor overall quality of the generated samples.

\subsection{Ablation Analysis}
In order to verify the effectiveness of DPO, as well as comparing with generating samples \textit{all at once}, we complete the ablation experiments for the following three different variants:

\paragraph{Generate all at once without DPO.} We change the key sentence in the Prompt for LLMs to \textit{``So please generate 32 samples for the relation `$\mathcal{R}$'. Please make the generated samples as different from the above demonstrations as possible.''}.

\paragraph{Generate one by one without DPO.} We change the key sentence in the Prompt for LLMs to \textit{``So please generate a sample for the relation `$\mathcal{R}$'. Please make the generated samples as different from the above demonstrations as possible.''}.

\paragraph{Generate one by one without DPO \& diversity instruction.} We change the key sentence in the Prompt for LLMs to \textit{``So please generate a sample for the relation `$\mathcal{R}$'.''}.

As shown in Table \ref{ablation}, in most cases, either in AAO or OBO generation mode, the performance of the generated training samples on the KnowPrompt mostly decrease after removing DPO fine-tuning, which indicates that reasonable DPO fine-tuning can help LLMs generate higher-quality training samples. Furthermore, it can be seen that the samples generated by OBO perform better on the RE models than those generated by AAO In most experimental settings, indicating that our proposed OBO generation mode is effective and can further improve the quality of training samples. Meanwhile, comparing the results of the experiments under the ``w/o DPO, AAO'' and ``w/o DPO, OBO'' conditions, it can be seen that the quality of the training samples under the former condition is higher. We think it is because in the ``OBO'' condition, the number of samples demonstrated in the Sample Demonstration Module in the Prompt increases, which reduces the attention to ``diversity instruction'', and makes the LLMs show ``Imitation Behavior'' when generating samples, resulting in higher similarity and lower quality of the final generated training samples. Finally, comparing the results of the experiments in the ``w/o DPO, OBO'' and ``w/o DPO, w/o DI, OBO'' conditions, it can be seen that the quality of the generated training samples is higher in the former condition, which suggests that the ``diversity instruction'' can remind the LLMs to take diversity into account when generating training samples. Also note that in the case of ``Re-TACRED, k=16'', the training samples generated by LLMs without DPO fine-tuning have poor diversity, but they perform better on Knowprompt model than the training samples generated by LLMs with DPO fine-tuning, which we think is related to the test data of Knowprompt model. The training samples generated by LLMs without DPO fine-tuning may just be close to the test data, so there is a phenomenon of low diversity but good training effect.

\begin{figure*}[h]
\centering
\centerline{\includegraphics[scale=0.198]{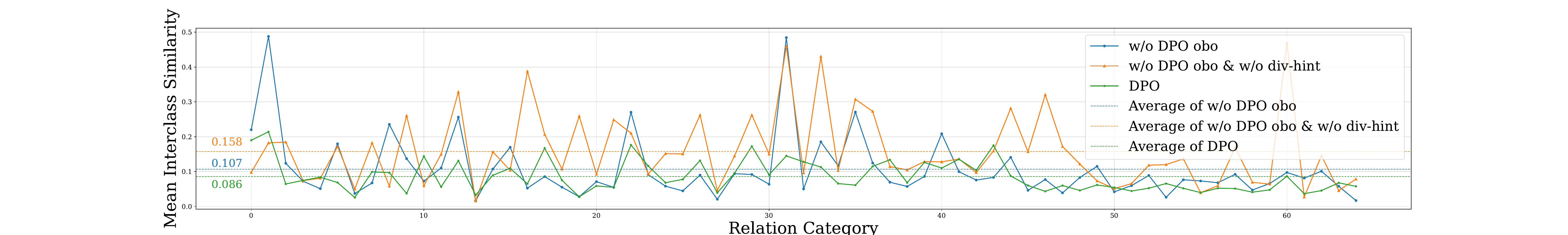}}
\caption{Average cosine similarity between generated training samples (K=32) for each relation category.}
\label{fig:diversity}
\end{figure*}

\begin{figure*}[h]
\centering
\centerline{\includegraphics[scale=0.198]{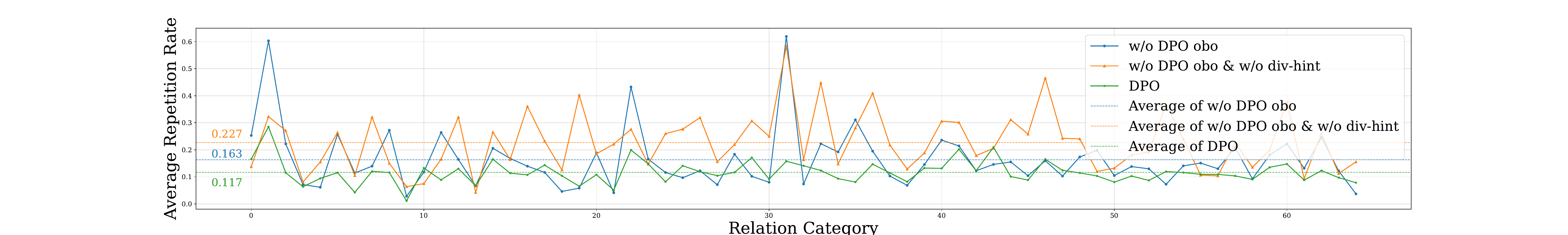}}
\caption{Average repetition rate of words between generated training samples (K=32) for each relation category.}
\label{fig:repetition}
\end{figure*}

\subsection{Diversity of Generated Samples}
\label{sub:diversity}
We extract the value of the "token" key from the sentence-level samples generated by LLMs, and then restore the words into sentences to facilitate similarity calculation. We calculate the average cosine similarity between every two pairs of generated training samples through OBO for each relation category, and the results are shown in Figure \ref{fig:diversity}. \textbf{The overall results show that the diversity of the relation extraction training samples generated by the LLMs after DPO fine-tuning training is lower than that of the two cases of LLMs w/o DPO}, which again illustrates the effectiveness of the DPO fine-tuning in the task of generating training samples. Furthermore, the instruction on ``diversity'' in the Prompt of the LLMs \textit{``Please make the generated samples as different from the above demonstrations as possible.''} also plays an important role, because the overall average diversity of the generated training samples is 0.051 lower than without this instruction. 

To further analyze the diversity, we also calculate the average repetition rate of words between every two pairs of the generated samples, as shown in Figure \ref{fig:repetition}. It is apparent that the repetition rate of words of training samples generated after DPO fine-tuning is lower than both of other cases.

\subsection{Number of Generated Samples}
\label{sub:number}
\begin{figure}[t]
\centering
\centerline{\includegraphics[scale=0.35]{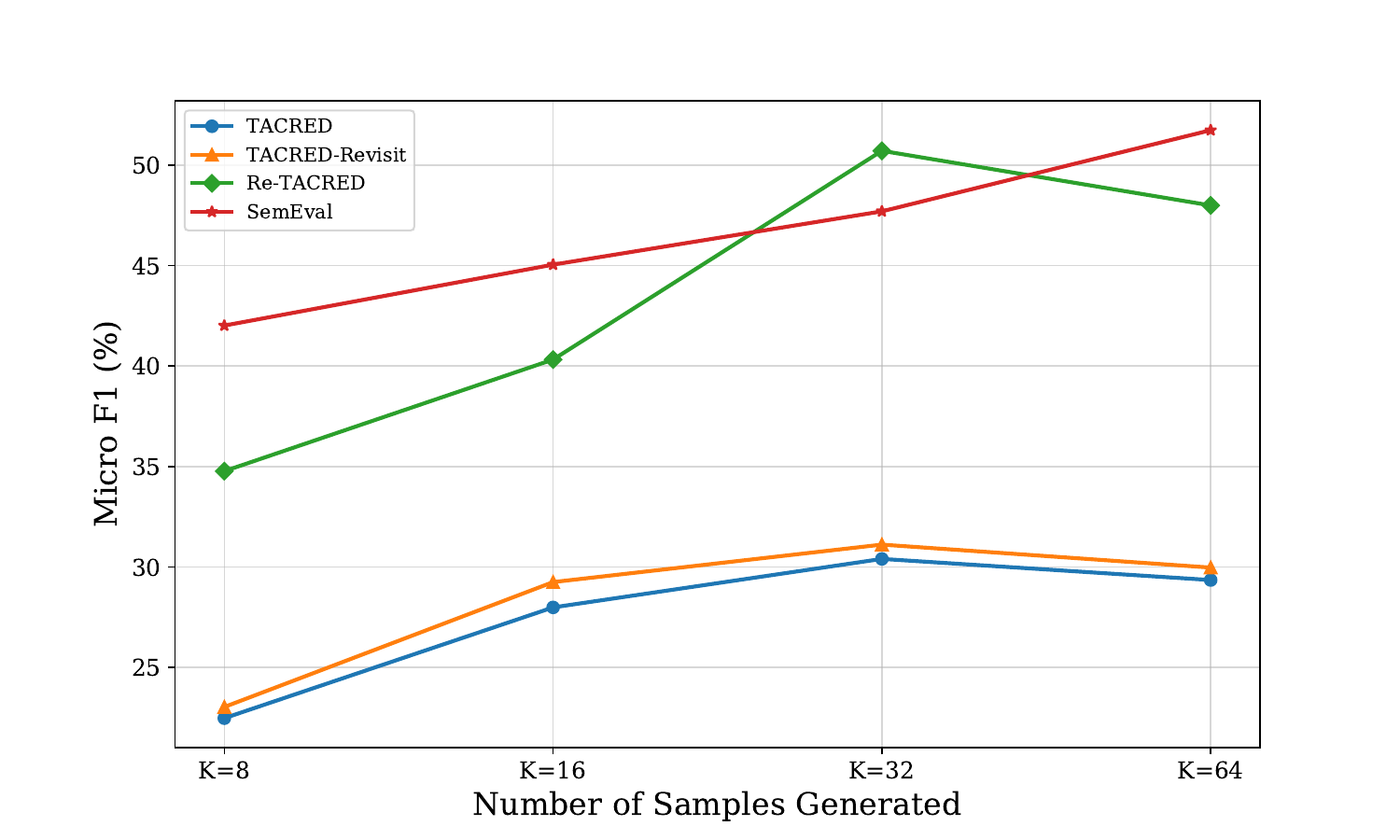}}
\caption{Micro F1 (\%) of KnowPrompt using different numbers of training samples generated by LLMs.}
\label{fig:number}
\end{figure}

\textbf{The generation of more training samples does not always improve the performance of the KnowPrompt, which has an upper limit.} We generated 8, 16, 32 and 64 samples on four relation extraction datasets using LLMs with ICL and DPO. From the results in Figure \ref{fig:number}, we find that the performance of the KnowPrompt increases and then remains constant as the number of samples generated by the LLMs increases, while peaking near K = 32. We argue that it is because the high-diversity of training samples generated by LLMs reaches the end point near K = 32 due to the limitation of the corpus training database of LLMs. It is also for this reason that the research in this paper is centered around 8-shot, 16-shot and 32-shot based.

\label{sub:Case}
\begin{table*}
\centering
\renewcommand{\arraystretch}{1.2}
\begin{tabularx}{\textwidth}{cc>{\centering\arraybackslash}X}
\hline
\textbf{Method} & \textbf{Step} & \textbf{Generated Data}\\
\hline
\multirow{8}*{Ours} & 1 & $\underline{\text{John Smith}}_{\text{subj}} \ \text{died in} \ \underline{\text{Canada}}_{\text{obj}} \ \text{after a long illness and was buried there.}$\\
\cmidrule(lr){2-3}
& 4 & $\underline{\text{Frank Ocean}}_{\text{subj}} \ \text{ended his life} \ \textbf{\textcolor{red}{in}} \ \underline{\text{France}}_{\text{obj}} \ \text{where he had lived for many years.} $\\
\cmidrule(lr){2-3}
& 6 & $\underline{\text{Bob Smith}}_{\text{subj}} \ \text{expired}  \ \textbf{\textcolor{red}{in}} \ \underline{\text{China}}_{\text{obj}}  \ \textbf{\textcolor{red}{where he}} \ \text{worked} \ \textbf{\textcolor{red}{for}} \ \text{decades.}$\\
\cmidrule(lr){2-3}
& 9 & $\underline{\text{Angela Davis}}_{\text{subj}} \ \text{passed away} \ \textbf{\textcolor{red}{in}} \ \underline{\text{Sweden}}_{\text{obj}} \ \textbf{\textcolor{red}{where}} \ \text{she fought} \ \textbf{\textcolor{red}{for}} \ \text{civil rights.}$ \\
\cmidrule(lr){2-3}
& 13 & $\underline{\text{James Bond}}_{\text{subj}} \ \textbf{\textcolor{red}{ended his life in}} \ \underline{\text{Monaco}}_{\text{obj}} \ \textbf{\textcolor{red}{where he}} \ \text{found himself unwanted.}$ \\
\cmidrule(lr){2-3}
& 16 & $\underline{\text{Tupac Shakur}}_{\text{subj}} \ \textbf{\textcolor{red}{passed away in}} \ \underline{\text{Grenada}}_{\text{obj}}. $\\
\hline
\multirow{8}*{\makecell[c]{Ours, \\ w/o DI, \\w/o DPO}} & 1 & $\underline{\text{France}}_{\text{obj}} \ \underline{\text{John F Kennedy}}_{\text{subj}} \ \text{American Kennedy family Assassination Dallas died.} $\\
\cmidrule(lr){2-3}
& 4 & $\underline{\text{China}}_{\text{obj}} \ \underline{\text{Wang Zhiyuan}}_{\text{subj}} \ \text{computer scientist} \ \textbf{\textcolor{red}{died}} \ \text{Beijing age 62.}$ \\
\cmidrule(lr){2-3}
& 6 & $\underline{\text{South Africa}}_{\text{obj}} \ \underline{\text{Nelson Mandela}}_{\text{subj}} \ \textbf{\textcolor{red}{died}} \ \text{Johannesburg} \ \textbf{\textcolor{red}{age}} \ \text{95.}$\\
\cmidrule(lr){2-3}
& 9 & $\underline{\text{Germany}}_{\text{obj}} \ \underline{\text{Heinrich Himmler}}_{\text{subj}} \ \text{Nazi official} \ \textbf{\textcolor{red}{died}} \ \text{Berlin} \ \textbf{\textcolor{red}{age}} \ \text{44.}$\\
\cmidrule(lr){2-3}
& 13 & $\textbf{\textcolor{red}{\text{China Wang Zhiyuan computer scientist died Beijing age 62.}}}$ \\
\cmidrule(lr){2-3}
& 16 & $\underline{\text{Canada}}_{\text{obj}} \ \underline{\text{Stephen Harper}}_{\text{subj}} \ \text{politician} \ \textbf{\textcolor{red}{died}} \ \text{Ottawa} \ \textbf{\textcolor{red}{age}} \ \text{68.}$ \\
\hline
\end{tabularx}
\caption{\label{tab:case study}
A case study of step-by-step generation of training samples on the ``per:country\_of\_death'' relation of TACRED. We mark in \textbf{\textcolor{red}{red}} the content of the training samples generated in the post-order that are identical to the content of the pre-order.
}
\end{table*}

\subsection{Case Study}
In response to the experimental phenomena in §\ref{sub:diversity} Diversity between Training Samples and §\ref{sub:number} Number of Generated Samples, we develop a specific case study using generated training samples on the ``per:country\_of\_death'' relation of TACRED. 

As shown in Table \ref{tab:case study}, our method allows LLMs to generate training samples with low similarity, and to maintain diversity in the verbalized representation of relations. On the contrary, after removing the ``diversity instruction'' and DPO, the generated training samples have higher similarity and single verbalized representation of relations, and even the post-order samples appear to be exactly the same as the pre-order samples. Meanwhile, we find that although our method enables LLMs to generate training samples with high diversity, the training samples generated by LLMs around Step = 16 also appear to be similar in the verbalized representations of relations, indicating that the similarity of the training samples generated later will increase. This results in the performance of the KnowPrompt peaking shortly after K = 16 and remaining largely unchanged thereafter.

\section{Conclusion}
In this paper, we propose a method for generating training samples for RE with LLMs. The method optimizes the output of LLMs to generate high-quality training samples for RE, especially in terms of diversity, through ICL and DPO. ICL allows LLMs to quickly learn the structure and content of training samples by providing appropriate sample demonstrations in the prompt; DPO allows LLMs to generate training samples with both diversity and correctness in mind through fine-tuning. Experiments demonstrate the effectiveness of these generated training samples in few-shot scenarios, especially with greater advantages in diversity. 



\section*{Limitations}
Despite our best efforts, the method proposed in this paper may still have some limitations.

\textbf{LLMs:} Although we have enabled the LLMs to generate better quality training samples by fine-tuning the training, the quality of these training samples is also largely limited by the strength of the open-source LLMs themselves.

\textbf{Maximum Number:} The maximum number of valid training samples that can be generated by LLMs is very limited, as performance does not consistently improve after generating about 16 or 32 training samples.

\section*{Acknowledgements}
This research is supported by the National Natural Science Foundation of China (No. 62306140, No. 62476127), the Natural Science Foundation of Jiangsu Province (No. BK20242039), the Basic Research Program of the Bureau of Science and Technology (ILF24001), the Fundamental Research Funds for the Central Universities (No. NJ2023032), the Scientific Research Starting Foundation of Nanjing University of Aeronautics and Astronautics (No. YQR21022), the Key Project of Jiangsu Collaborative Innovation Center of Chinese Medicinal Resources Industrialization (No. 000003401025-6), the Open Project of Chinese Materia Medica First-Class Discipline of Nanjing University of Chinese Medicine (No. ZYXJC2024-010) and the High Performance Computing Platform of Nanjing University of Aeronautics and Astronautics.

\bibliography{anthology,custom}
\bibliographystyle{acl_natbib}

\appendix

\section{Experimental Details}
\label{sec:appendix}
\subsection{Datasets}
For comprehensive experiments, we conducted experiments on four relation extraction datasets: TACRED, TACRED-Revisit, Re-TACRED and SemEval 2010 Task 8 (SemEval). A brief introduction to these data is given below:

\textbf{TACRED:} a large-scale sentence-level relation extraction dataset from the annual TACBP4 challenge, containing over 106,000 sentences. It involves 42 different relation categories, including 41 common relation categories and a special ``no relation'' relation category.

\textbf{TACRED-Revisit:} a dataset constructed on the basis of the TACRED dataset. The researchers found errors in the development and test sets of the original TACRED dataset and corrected them while keeping the training set intact.

\textbf{Re-TACRED:} another version of the TACRED dataset, which addresses some of the shortcomings of the original TACRED dataset by reconstructing the training, validation and test sets. Meanwhile, this dataset removes the original 6 relation categories and adds 4 new relation categories to the TACRED dataset, so that a dataset with 40 relation categories is finally obtained.

\textbf{SemEval:} a traditional relation extraction dataset, containing 10,717 annotated samples, covers 9 bi-directional relation categories and a special ``no relation'' relationship category.
\label{sub:datasets}



\subsection{Relation Explanation}
\label{sub:RES}
We give explanations for each relation in the four datasets. The detailed explanation for each relation is shown in Table \ref{tab:Relation Explanation}.

\begin{table*}
\centering
\renewcommand{\arraystretch}{1.2}
\begin{tabularx}{\textwidth}{l|>{\arraybackslash}X}
    \hline
    \textbf{Relation} & \textbf{Explanation}\\
    \hline
    Component-Whole (e2,e1)& Tail entity e2 is the component of head entity e1, and head entity e1 is the whole of tail entity e2 \\
    \hline
    Instrument-Agency (e2,e1)	&Tail entity e2 is the instrument of head entity e1, and head entity e1 is the agency of tail entity e2\\
    \hline
    Member-Collection (e1,e2)	&Head entity e1 is the member of tail entity e2, and tail entity e2 is the collection of head entity e1\\
    \hline
    Cause-Effect (e2,e1)&	Tail entity e2 is the cause of head entity e1, and head entity e1 is the effect of tail entity e2\\
    \hline
    Entity-Destination (e1,e2)&	Head entity e1 is the entity of tail entity e2, and tail entity e2 is the destination of head entity e1\\
    \hline
    Content-Container (e1,e2)	&Head entity e1 is the content of tail entity e2, and tail entity e2 is the container of head entity e1\\
    \hline
    Message-Topic (e1,e2)&	Head entity e1 is the message of tail entity e2, and tail entity e2 is the topic of head entity e1\\
    \hline
    Product-Producer (e2,e1)&	Tail entity e2 is the product of head entity e1, and head entity e1 is the producer of tail entity e2\\
    \hline
    Member-Collection (e2,e1)&	Tail entity e2 is the member of head entity e1, and head entity e1 is the collection of tail entity e2\\
    \hline
    Entity-Origin (e1,e2)	&Head entity e1 is the entity of tail entity e2, and tail entity e2 is the origin of head entity e1\\
    \hline
    Cause-Effect (e1,e2)	&Head entity e1 is the cause of tail entity e2, and tail entity e2 is the effect of head entity e1\\
    \hline
    Component-Whole (e1,e2)&	Head entity e1 is the component of tail entity e2, and tail entity e2 is the whole of head entity e1\\
    \hline
    Message-Topic (e2,e1)&	Tail entity e2 is the message of head entity e1, and head entity e1 is the topic of tail entity e2\\
    \hline
    Product-Producer (e1,e2)&	Head entity e1 is the product of tail entity e2, and tail entity e2 is the producer of head entity e1\\
    \hline
    Entity-Origin (e2,e1)	&Tail entity e2 is the entity of head entity e1, and head entity e1 is the origin of tail entity e2\\
    \hline
    Content-Container (e2,e1)&	Tail entity e2 is the content of head entity e1, and head entity e1 is the container of tail entity e2\\
    \hline
    Instrument-Agency (e1,e2)&	Head entity e1 is the instrument of tail entity e2, and tail entity e2 is the agency of head entity e1\\
    \hline
    Entity-Destination (e2,e1)	&Tail entity e2 is the entity of head entity e1, and head entity e1 is the destination of tail entity e2\\
    \hline
    Other	&Tail entity e2 is the component of head entity e1, and head entity e1 is the whole of tail entity e2\\
    \hline
    org:founded&	The founding relationship of an organization\\
    \hline
    org:subsidiaries	&The subsidiaries of an organization\\
    \hline
    per:date\_of\_birth&	The date of birth of a person\\
    \hline
    per:cause\_of\_death	&The cause of death of a person\\
    \hline
    per:age	&The age of a person\\
    \hline
    per:stateorprovince\_of\_birth&	The state or province of birth of a person\\
    \hline
    per:countries\_of\_residence	&The countries where a person resides\\
    \hline
\end{tabularx}
\end{table*}

\begin{table*}
\centering
\renewcommand{\arraystretch}{1.2}
\begin{tabularx}{\textwidth}{l|>{\arraybackslash}X}
    \hline
    \textbf{Relation} & \textbf{Explanation}\\
    
    \hline
    per:country\_of\_birth&	The country of birth of a person\\
    \hline
    per:stateorprovinces\_of\_residence	&The states or provinces where a person resides\\
    \hline
    org:website&	The website of an organization\\
    \hline
    per:cities\_of\_residence&	The cities where a person resides\\
    \hline
    per:parents	&The parents of a person\\
    \hline
    per:employee\_of	&The organization where a person is employed\\
    \hline
    NA/no\_relation	&Unknown relation\\
    \hline
    per:city\_of\_birth	&The city of birth of a person\\
    \hline
    org:parents	&The parent company of an organization\\
    \hline
    org:political/religious\_affiliation&	The political or religious affiliation of an organization\\
    \hline
    per:schools\_attended&	The schools attended by a person\\
    \hline
    per:country\_of\_death	&The country where a person died\\
    \hline
    per:children&	The children of a person\\
    \hline
    org:top\_members/employees&	The top members/employees of an organization\\
    \hline
    per:date\_of\_death	&The date of death of a person\\
    \hline
    org:members&	The members of an organization\\
    \hline
    org:alternate\_names&	The alternate names of an organization\\
    \hline
    per:religion	&The religion of a person\\
    \hline
    org:member\_of&	The organization to which a member belongs\\
    \hline
    org:city\_of\_headquarters	&The city where the headquarters of an organization is located\\
    \hline
    per:origin	&The origin of a person\\
    \hline
    org:shareholders	&The shareholders of an organization\\
    \hline
    per:charges&	The charges against a person\\
    \hline
    per:title&	The title of a person\\
    \hline
    org:number\_of\_employees/members&	The number of employees/members in an organization\\
    \hline
    org:dissolved	&The date of dissolution of the organization\\
    \hline
    org:country\_of\_headquarters	&The country where headquarters of an organization is located\\
    \hline
    per:alternate\_names	&The alternate names of a person\\
    \hline
    per:siblings	&The siblings of a person\\
    \hline
    org:stateorprovince\_of\_headquarters&	The state or province where headquarters of an organization is located\\
    \hline
    per:spouse	&The spouse of a person\\
    \hline
    per:other\_family	&Other family members of a person\\
    \hline
    per:city\_of\_death	&The city where a person died\\
    \hline
    per:stateorprovince\_of\_death&	The state or province where a person died\\
    \hline
    org:founded\_by&	The founder of an organization\\
    \hline
    org:country\_of\_branch&	The country where a branch of an organization is located\\
    \hline
    org:city\_of\_branch	&The city where a branch of an organization is located\\
    \hline
    org:stateorprovince\_of\_branch&	The state or province where branch of an organization is located\\
    \hline
    per:identity	&The identity information or characteristics of a person\\
    \hline
\end{tabularx}
\caption{\label{tab:Relation Explanation}
Explanation of each relation in the four datasets.  
}
\end{table*}

\end{document}